\documentclass[%
twocolumn,
superscriptaddress,
 amsmath,amssymb,
 aps,
]{revtex4-1}

\usepackage{graphicx}% Include figure files
\usepackage{booktabs}
\usepackage{bm}% bold math
\usepackage{color}
\definecolor{bluecolor}{rgb}{0,0.,1.}

\definecolor{redcolor}{rgb}{.7,0.,0.}

\begin{document}

\title{A standardized Project Gutenberg corpus for statistical analysis of natural language and quantitative linguistics }

\author{Martin Gerlach}
\email{martin.gerlach@northwestern.edu}
\affiliation{Department of Chemical and Biological Engineering, Northwestern University, Evanston, IL 60208, USA}

\author{Francesc Font-Clos}
\email{francesc.font@unimi.it}
\affiliation{Center for Complexity and Biosystems, Department of Physics,\\University of Milan, 20133 Milano, Italy}


\begin{abstract}
The use of Project Gutenberg (PG) as a text corpus has been extremely popular in statistical analysis of language for more than 25 years.
However, in contrast to other major linguistic datasets of similar importance, no consensual full version of PG exists to date.
In fact, most PG studies so far either consider only a small number of manually selected books, leading to potential biased subsets, 
or employ vastly different pre-processing strategies (often specified in insufficient details), raising concerns regarding the reproducibility of published results.
In order to address these shortcomings, here we present the Standardized Project Gutenberg Corpus (SPGC), an open science approach to a curated version of the complete PG data containing more than 50,000 books and more than $3 \times 10^9$ word-tokens.
Using different sources of annotated metadata, we not only provide a broad characterization of the content of PG, but also show different examples highlighting the potential of SPGC for investigating language variability across time, subjects, and authors.
We publish our methodology in detail, the code to download and process the data, as well as the obtained corpus itself on 3 different levels of granularity (raw text, timeseries of word tokens, and counts of words).
In this way, we provide a reproducible, pre-processed, full-size version of Project Gutenberg as a new scientific resource for corpus linguistics, natural language processing, and information retrieval.

%By publishing the methodology and the code, as well as the fully processed dataset on 3 different levels of granularity (raw text, timeseries of word tokens, and counts of words) this resource allows for easier access to the usage of PG for statistical analysis of language and quantitative linguistics; as well as a necessary step towards reproducibility across different studies in the framework of open science.
% \Red{Use this defaults: https://gist.github.com/fontclos/a398c95d2841dfb19e864d08c1f60201}
\end{abstract}

\maketitle

% \section{Notes of progress}
% 
% Embedding
% 
% A figure with statistics

\section{Introduction}

% Analysis of natural  language from a complex systems perspective has provided new insights into statistical properties of language~\cite{Gerlach2018-hv, Lippi2018-ua, Cocho2015-hu, Sole2010-hi, Gerlach2016-ld, Gerlach2013-jf, Font-Clos2015-qy, Ferrer_i_Cancho2003-he, Corominas-Murtra2015-mp, Amato2018-cu}. \Red{for example ...?}
Analysis of natural  language from a complex systems perspective has provided new insights into statistical properties of language, such as statistical laws~\cite{Altmann2016-nl,Ferrer_i_Cancho2003-he,Petersen2012-of,Tria2014-li, Corominas-Murtra2015-mp, Font-Clos2015-qy, Cocho2015-hu, Lippi2018-ua,Mazzolini2018-po},
networks~\cite{Dorogovtsev2001-yi,Sole2010-hi,Amancio2011-ww,Choudhury2010-bk,Cong2014-ia},
language change~\cite{Bochkarev2014-kb, Ghanbarnejad2014-qx,Feltgen2017-wy,Goncalves2018-yf,Amato2018-cu,Karjus2018-ne}, 
quantification of information content~\cite{Montemurro2010-mi,Takahira2016-mn,Febres2017-nt,Bentz2017-ux}, 
or the role of syntactic structures~\cite{Ferrer_i_Cancho2004-eu} or punctuation~\cite{Kulig2016-hy}, etc.
In particular, the availability of new and large publicly available datasets such as the google-ngram data~\cite{Michel2011-ya}, the full Wikipedia dataset~\cite{Masucci2011-db, Yasseri2012-lf}, or Twitter~\cite{Dodds2011-je} opened the door for new large-scale quantitative approaches.
% While these databases are unprecedented in size containing millions of documents, they suffer from a variety of drawbacks: their composition is unknown (e.g. google-ngram data, see~\cite{Morse-Gagne2011-pj, Pecheniack2015}), or they contain mostly very short texts (e.g. Use-net groups ~\cite{Altmann2011} or Twitter~\cite{?}).
One of the main drawbacks of these datasets, however, is the lack of ``purity'' of the samples resulting from the fact that (i) the composition of the dataset is largely unknown (google-ngram data, see~\cite{Morse-Gagne2011-pj, Pechenick2015-ov}), (ii) the texts are a mixture of different authors (Wikipedia), or (iii) the texts are extremely short (Twitter).
% One way to address this issue is to combine different texts to obtain longer texts, at the expense of mixing different authors and genres. The effect of mixing texts is not well-understood , and it has been claimed~\cite{Williams2015-mz} that some well-known statistical features of large aggregated corpora might be a byproduct of the aggregation process itself. 
One approach to ensure large homogeneous samples of data is to analyze literary books -- the most popular being from Project Gutenberg (PG)~\cite{Hart_undated-vb} due to their free availability.

Data from PG has been used in numerous cases to quantify statistical properties of natural language.
In fact, the statistical analysis of texts in the framework of complex systems or quantitative linguistics is not conceivable without the books from PG.
Already in the 1990's the seminal works by Ebeling et al.~\cite{Ebeling1994-nc} and Schurman \& Grassberger ~\cite{Schurmann1996-sv} used up to 100 books from PG in the study of long-range correlations and Baayen~\cite{Baayen1996-oc} investigated the growth curve of the vocabulary.
Subsequently, PG has become an indispensable resource for the quantitative analysis of language investigating, e.g., universal properties (such as correlations~\cite{Altmann2012-bk} or scale-free nature of the word-frequency distribution~\cite{Moreno-Sanchez2016-ty, Williams2015-mz, Tria2018-ic} ) or aspects related to genres~\cite{Hughes2012-xs} or emotions~\cite{Reagan2016-yp}.

%However, the current practice of using data from PG raises two major concerns in terms of its scientific rigor \Red{PERHAPS TOO STRONG}.
%
While we acknowledge that PG has so far been of great use to the community, we also find that it has been handled in a careless and unsystematic way. Our criticisms can be summarized in two points.
First, the majority of studies only consider a small subset (typically not more than 20 books) from the more than 50,000 books available in PG.
More importantly, the subsets often contain the same manually selected books such as the work ``Moby Dick'' which can be found in virtually any study using PG data.
Thus different works potentially employ biased and correlated subsets.
% raising questions about the generalizibilty of the respective findings.
% Most of the studies used fewer than 100 books; mostly a biased sample.
% While these studies have allowed us to gain deeper insights into the structure of natural language, the use of this data suffers from severe limitations.
% First, typically these studies consider only a manually selected subset from the more than 50,000 available books.
% For example, from XX papers using not more than 20 books, we found Moby Dick X times, ...
% Possible reasons for this are the lack of easy accessiblity of the data and the requirement to filter headers and spurious information contained in each book.
% As a result, the subsets are not only small but also correlated and potentially biased.
% Aren't there cases in sequencing/... that anlyzed particular model organisms that are not representative of general mechanisms?
% Only few exceptions have tried to analyse larger or full samples -- see the list. 
% However, description of mining and processing steps leave out many details; neither have code or the final data been published.
%
Second, different studies use different filtering techniques to mine, parse, select, tokenize, and clean the data or do not describe the methodological steps in sufficient detail.
As a result, two studies using the supposedly same PG data might end up with somewhat different datasets.

Taken together, these limitations raise concerns about the replicability and generalizability of previous and future studies.
In order to ensure the latter, it is pertinent to make corpora widely available in a standardized format. 
While this has been done for many textual datasets in machine learning (e.g. the UCI machine learning repository~\cite{Dua:2017}) and diachronic corpora for studying language change (e.g. The Corpus of Contemporary American English~\cite{COCA-corpus}), such efforts have so far been absent for data from PG.

Here, we address these issues by presenting a standardized version of the complete Project Gutenberg data --- the Standardized Project Gutenberg Corpus (SPGC) --- containing more than 50,000 books and more than $3 \times 10^9$ word-tokens.
We provide a framework to automatically download, filter, and process the raw data on three different levels of granularity: i) the raw text, ii) a filtered timeseries of word-tokens, and iii) a list of occurrences of words.
We further annotate each book with metadata about language, author (name and year of birth/death), and genre as provided by Project Gutenberg records as well as through collaborative tagging (so-called bookshelves), and show that the latter has more desirable properties such as low overlap between categories.
We exemplify the potential of the SPGC by studying its variability in terms of Jensen-Shannon divergence across authors, time and genres.

In contrast to standard corpora such as the British National Corpus~\cite{Leech1993-ub} or the Corpus of Contemporary American English~\cite{COCA-corpus}, the new Standardized Project Gutenberg Corpus is decentralized, dynamic and multi-lingual.
The SPGC is decentralized in the sense that anyone can recreate it from scratch in their computer executing a simple python script.
The SPGC is dynamic in the sense that, as new books are added to PG, the SPGC incorporates them immediately, and users can update their local copies with ease.
This removes the classic centralized dependency problem, where a resource is initially generated by an individual or institution and initially maintained for certain period of time, after which the resource is not updated anymore and remains ``frozen'' in the past. 
Finally, the SPGC is multi-lingual because it is not restricted to any language, it simply incorporates all content available in PG (see Section \ref{sec:data_description} for details).
Thus, in order to be compatible with a standard corpus model, and to ensure reproducibility of our results, we also provide a static time-stamped version of the corpus, SPGC-2018-07-18 (\href{https://doi.org/10.5281/zenodo.2422560}{https://doi.org/10.5281/zenodo.2422560}).

In summary, we hope that the SPGC will lead to an increase in the availability and reliability of the PG data in the statistical and quantitative analysis of language.

\section{Data acquisition}
Project Gutenberg is a digital library founded in 1971 which archives cultural works uploaded by volunteers.
The collection primarily consists of copyright-free literary works (books), currently more than 50,000, and is intended as a resource for readers to enjoy literary works that have entered the public domain.
Thus the simplest way for a user to interact with PG is through its website, which provides a search interface, category listings, etc. to facilitate locating particular books of interest. 
Users can then read them online for free, or download them as plain text or ebook format.
While such a manual strategy might suffice to download tens or hundreds of books (given the patience) it does not reasonably scale to the complete size of the PG data with more than 50,000 books.

Our approach consists of downloading the full PG data automatically through a local mirror, see \href{https://www.gutenberg.org/wiki/Gutenberg:Information_About_Robot_Access_to_our_Pages}{Project Gutenberg's Information About Robot Access page} for details. We keep most technical details ``under the hood'' and instead present a simple, well structured solution to acquire all of PG with a single command.

In addition to the book's data, our pipeline automatically retrieves two different datasets containing annotations about PG books. 
The first set of metadata is provided by the person who uploads the book, and contains information about the author (name, year of birth, year of death), language of the text, subject categories, and number of downloads.
The second set of metadata, the so-called bookshelves, provide a categorization of books into collections such as ``Art'' or ``Fantasy'', in analogy to the process of collaborative tagging~\cite{Cattuto2007-uo}.

\begin{figure}
 \centering
 \includegraphics[width=\columnwidth]{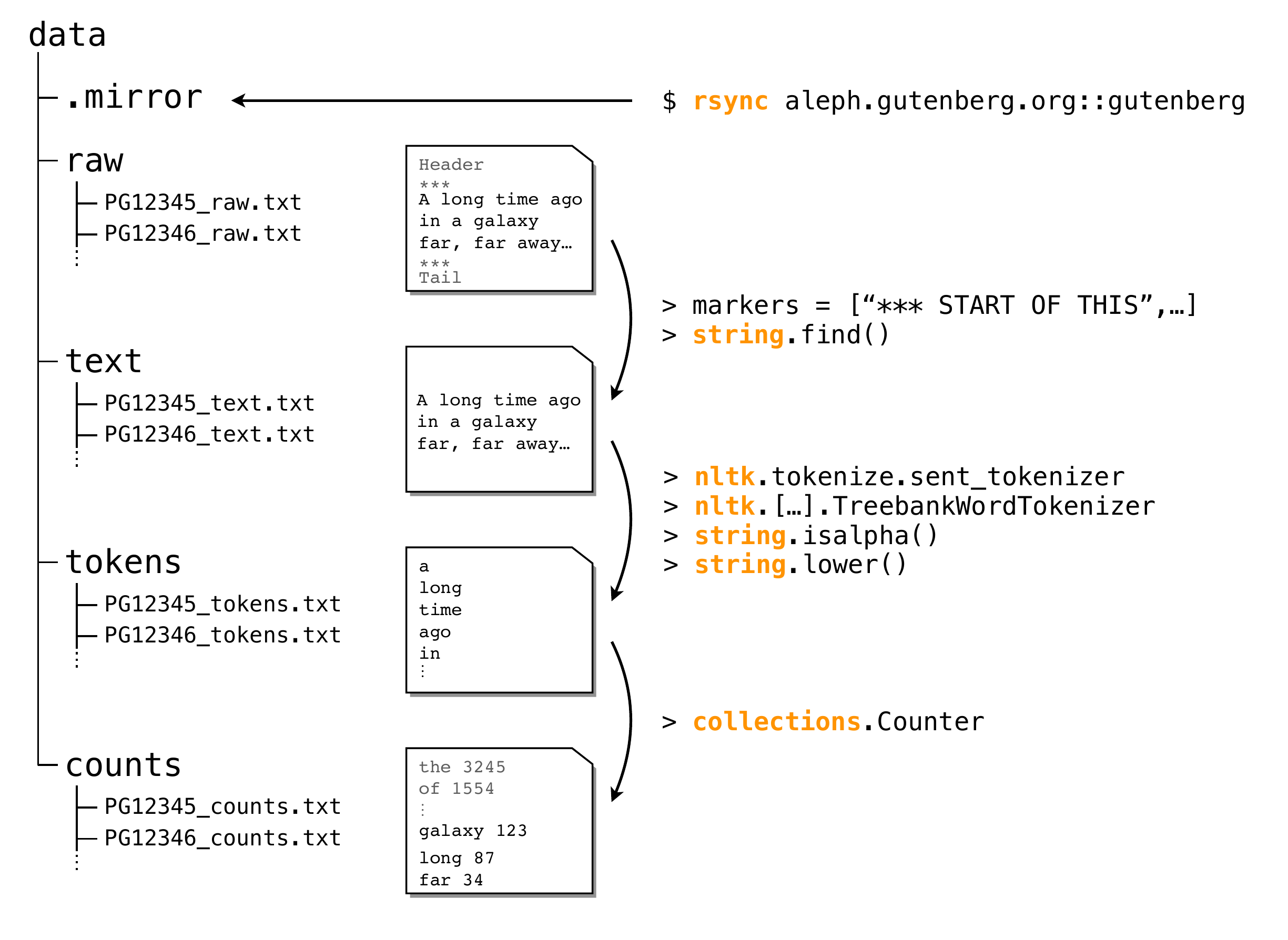}
 \caption{Sketch of the pre-processing pipeline of the PG data. The folder structure (left) organizes each PG book on 4 different levels of granularity, see example books (middle): raw, text, tokens, and counts.
 On the right we show the basic python commands used in the pre-processing.}
 \label{fig.pipeline}
\end{figure}
\section{Data processing}
In this section we briefly describe all steps we took to obtain the corpus from the raw data (Fig. \ref{fig.pipeline}), for details see Sec.~\ref{sec.mat.met}.
The processing (as of July 18, 2018) yields data for 55,905 books on 4 different levels of granularity: 

\begin{itemize}
 \item \textit{Raw} data: We download all books and save them according to their PG-ID. We eliminate duplicated entries and entries not in UTF-8 encoding.

 \item \textit{Text} data: We remove all headers and boiler plate text, see Methods for details.
 
 \item \textit{Token} data: We tokenize the text data using the tokenizer from NLTK~\cite{Loper2002-vq}. This yields a time series of tokens without punctuation, etc. 
  
 \item \textit{Count} data: We count the number of occurrences of each word-type. This yields a list of tuples ($w$,$n_w$), where $w$ is word-type $w$ and $n_w$ is the number of occurrences.

\end{itemize}

\section{Data Description}
\label{sec:data_description}
\begin{figure}
 \centering
 \includegraphics[width=\columnwidth]{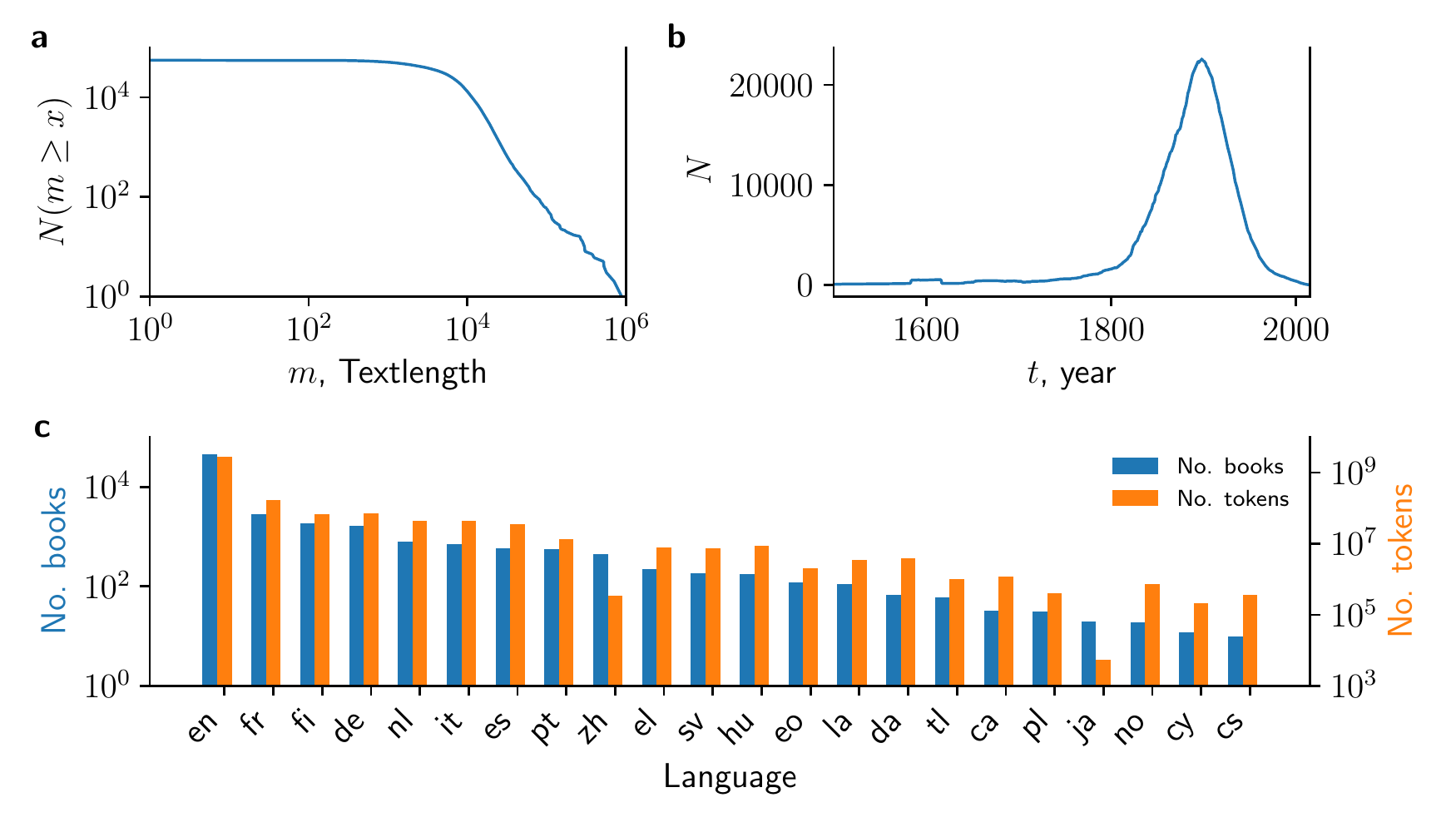}
 \caption{Basic summary statistics from the processed PG data. 
 (a) Number of books with a text length larger than $m$.
 (b) Number of books which are compatible with being published in year $t$, i.e. year of author's birth is 20 years prior and year of author's death is after $t$.
 (c) Number of books (left axis) and number of tokens (right axis) which are assigned to a given language based on the metadata. en: English, fr: French, fi: Finnish, de: German, nl: Dutch, it: Italian, es: Spanish, pt: Portuguese, zh: Chinese, el: Greek, sv: Swedish, hu: Hungarian, eo: Esperanto, la: Latin, da: Danish, tl: Tagalog, ca: Catalan, pl: Polish, ja: Japanese, no: Norwegian, cy: Welsh, cs: Czech.}
 \label{fig.stats_summary}
\end{figure}

We provide a broad characterization of the PG books in terms of their length, language and (when available) inferred date of publication in Fig.~\ref{fig.stats_summary}.

One of the main reason for the popularity of books from PG is their long text length, which yields large coherent statistical samples without potentially introducing confounding factors originating from, e.g., the mixing of different texts~\cite{Williams2015-mz}.
The length of most PG books exceeds $m=10^4$ word tokens (Fig.~\ref{fig.stats_summary}a) larger than typical documents from most web-resources.
In fact, the distribution shows a heavy-tail for large values of $m$.
Thus we find a substantial fraction of books having more than $10^5$ word tokens.

Many recent applications in quantitative linguistic aim at tracing diachronic changes.
While the metadata does not provide the year of the first publication of each book, we approximate the number of PG books published in year $t$ as the number of PG books for which the author's year of birth is $t_\mathrm{birth} + 20 < t$ and the author's year of death is $t < t_\mathrm{death}$ (Fig.~\ref{fig.stats_summary}b).
This reveals that the vast majority of books were first published around the year 1900, however, with a substantial number of books between 1800 and 2000. Part of this is known to be a consequence of the Copyright Term Extension Act of 1998 which, sadly, has prevented books published after 1923 to enter the public domain so far. If no further copyright extensions laws are passed in the future, then this situation will be gradually alleviated year after year, as books published in 1923 will enter the public domain on January 1st, 2019, and so on.

While most contemporary textual datasets are in English, the SPGC provides a rich resource to study other languages.
Using metadata provided by PG, we find that 81\% of the books are tagged as written in English, followed by French (5\%, 2864 books), Finnish (3.3\%, 1903 books) and German (2.8\%, 1644 books). In total, we find books written in 56 different languages, with 3 (13) languages besides English with more than $1,000$ ($100$) books each (Fig.~\ref{fig.stats_summary}c). 
The size of the English corpus is $2.8 \times 10^9$ tokens, which is more than one order of magnitude larger than the British National Corpus ($10^8$ tokens).
The second-largest language corpus is made up of French books with $>10^8$ tokens.
Notably, there are six other languages (Finnish, German, Dutch, Italian, Spanish, and Portuguese) that contain $>10^7$ tokens and still another $8$ languages (Greek, Swedish, Hungarian, Esperanto, Latin, Danish, Tagalog, and Catalan) that contain $>10^6$ tokens.

\begin{figure}
 \centering
 \includegraphics[width=\columnwidth]{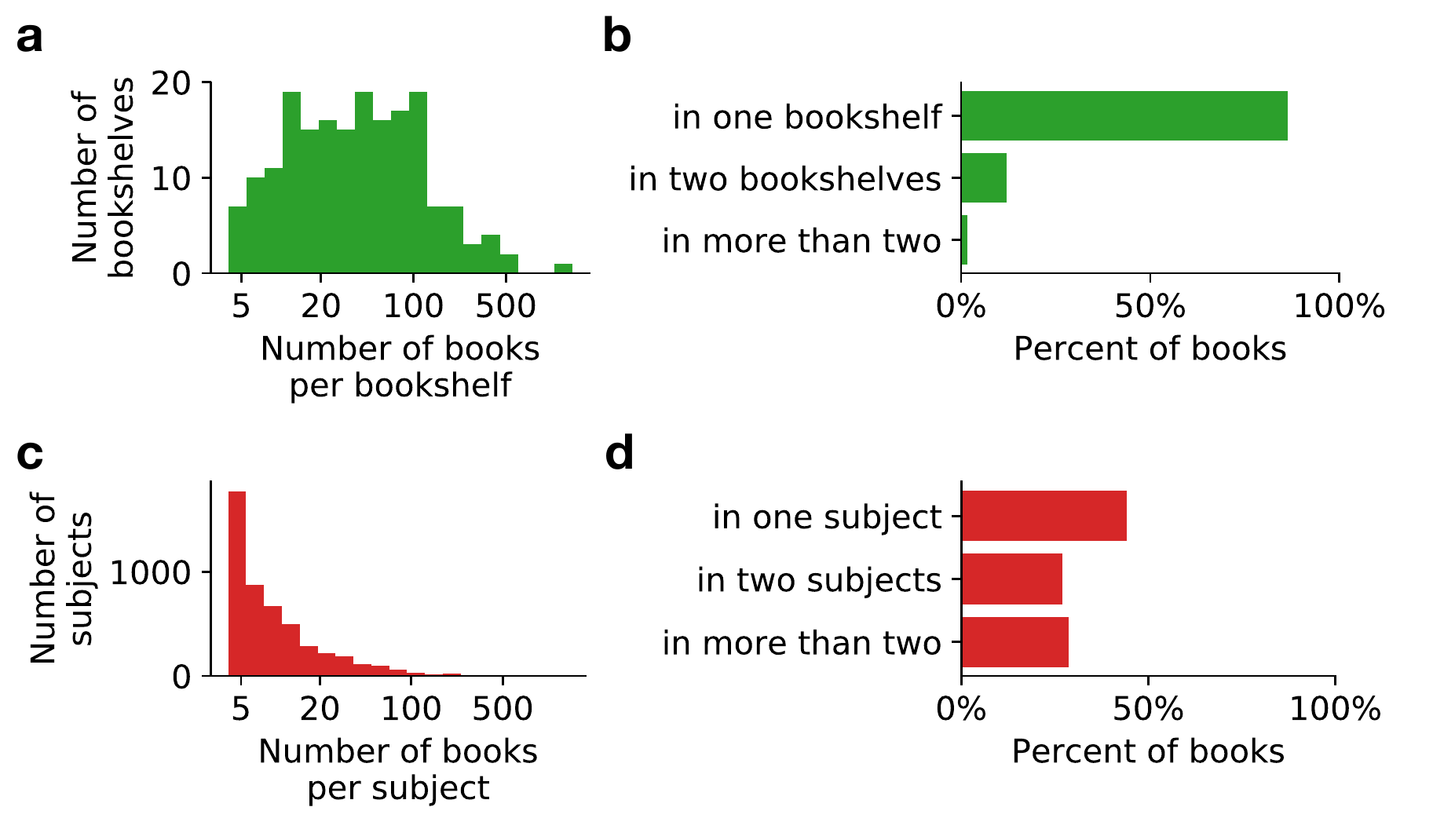}
 \caption{Comparison between bookshelf labels (top, green) and subject labels (bottom, red).
 (a,c) Number of labels with a given number of books.
 (b,d) Fraction of books with a given number of labels.}
 \label{fig.stats_subjects}
\end{figure}

In addition to the ``hard-facts'' metadata (such as language, time of publication), the SPGC also contains manually annotated topical labels for individual books.
These labels allow not only the study of topical variability, but they are also of practical importance for assessing the quality of machine learning applications in Information Retrieval, such as text classification or topic modeling \cite{Manning2008-sf}.
We consider two sets of topical labels: labels obtained from PG's metadata ``subject'' field, which we call \emph{subject labels};
and labels obtained by parsing PG's website bookshelf pages, which we call \emph{bookshelf labels}.
Table~\ref{tab:labels} shows that there is certain overlap in the most common labels between the two sets (e.g. Science Fiction or Historical Fiction), but a more detailed analysis of how labels are assigned to books reveals substantial differences (Fig.~\ref{fig.stats_subjects}).
First, subject labels display a very uneven distribution of the number of books per label. 
That is, most of the subject labels are assigned to very few books (less than 10), with only few subject labels assigned to many books. 
In comparison, bookshelf labels are more evenly distributed: most of them are assigned to between 10 and 100 books (Fig.~\ref{fig.stats_subjects}a,c).
More importantly, the overlap in the assignment of labels to individual books is much smaller for the bookshelf labels (Fig.~\ref{fig.stats_subjects}b,d): While roughly 50\% of the PG books are tagged with two ore more subject labels, up to 85\% of books are tagged with a unique bookshelf label.
This indicates that the bookshelf labels are more informative because they constitute broader categories and provide a unique assignment of labels to books.
Thus, our analysis suggests that these labels are better suited for practical applications such as text classification. 

\begin{table}
\resizebox{1\columnwidth}{!}{  
\begin{tabular}{rrl}
\hline
\hline
\toprule 
Rank & Books &                             Bookshelf \\
\midrule
\hline
1    &            1341 &                       Science Fiction \\
2    &             509 &                Children's Book Series \\
3    &             493 &                                 Punch \\
4    &             426 &      Bestsellers, American, 1895-1923 \\
5    &             383 &                    Historical Fiction \\
6    &             374 &                           World War I \\
7    &             339 &                    Children's Fiction \\
...  &             ... &                                   ... \\
47   &              94 &                               Slavery \\
48   &              92 &                               Western \\
49   &              90 &                               Judaism \\
50   &              86 &                   Scientific American \\
51   &              84 &   Pirates, Buccaneers, Corsairs, etc. \\
52   &              83 &                    Astounding Stories \\
53   &              83 &                 Harper's Young People \\
...  &             ... &                                   ... \\
97   &              37 &  Animals-Wild-Reptiles and Amphibians \\
98   &              37 &                         Short Stories \\
99   &              36 &                   Continental Monthly \\
100  &              35 &                          Architecture \\
101  &              35 &                          Bahá'í Faith \\
102  &              34 &         Precursors of Science Fiction \\
103  &              33 &                               Physics \\
...  &             ... &                                   ... \\
\bottomrule
\hline
\hline
\end{tabular}
% \label{tab:bookshelves}
% \caption{Examples of bookshelves names}
% \end{table}
% 
% \begin{table}
\begin{tabular}{rrl}
\hline
\hline
\toprule
Rank & Books &                             Subject \\
\hline
\midrule
1    &  2006 &                             Fiction \\
2    &  1823 &                       Short stories \\
3    &  1647 &                     Science fiction \\
4    &   913 &                   Adventure stories \\
5    &   746 &                  Historical fiction \\
6    &   708 &                        Love stories \\
7    &   690 &                              Poetry \\
...  &   ... &                                 ... \\
47   &   190 &             Short stories, American \\
48   &   188 &              Science -- Periodicals \\
49   &   183 &                     American poetry \\
50   &   180 &                               Drama \\
51   &   165 &           Paris (France) -- Fiction \\
52   &   163 &                  Fantasy literature \\
53   &   162 &                  Orphans -- Fiction \\
...  &   ... &                                 ... \\
97   &   100 &             Scotland -- Periodicals \\
98   &    98 &                        Horror tales \\
99   &    97 &                   Canada -- Fiction \\
100  &    97 &       France -- Court and courtiers \\
101  &    96 &           Social classes -- Fiction \\
102  &    95 &                Courtship -- Fiction \\
103  &    95 &  Seafaring life -- Juvenile fiction \\
...  &   ... &                                 ... \\
\bottomrule
\hline
\hline
\end{tabular}
} % from resizebox0command
\caption{Examples for the names of labels and the number of assigned books from Bookshelves (left) and Subjects (right) metadata.}
\label{tab:labels}
\end{table}

\section{Quantifying variability in the corpus}
In order to highlight the potential of the SPGC for quantitative analysis of language, we quantify the degree of variability in the statistics of word frequencies across labels, authors, and time.
For this, we measure the distance between books $i$ and $j$ using the well-known Jensen-Shannon divergence~\cite{Gerlach2016-ld}, $D_{i,j}$, with $D_{i,j}=0$ if the two books are exactly equal in terms of frequencies, and $D_{i,j}=1$ if they are maximally different, i.e. they do not have a single word in common.

\begin{figure}
 \centering
 \includegraphics[width=\columnwidth]{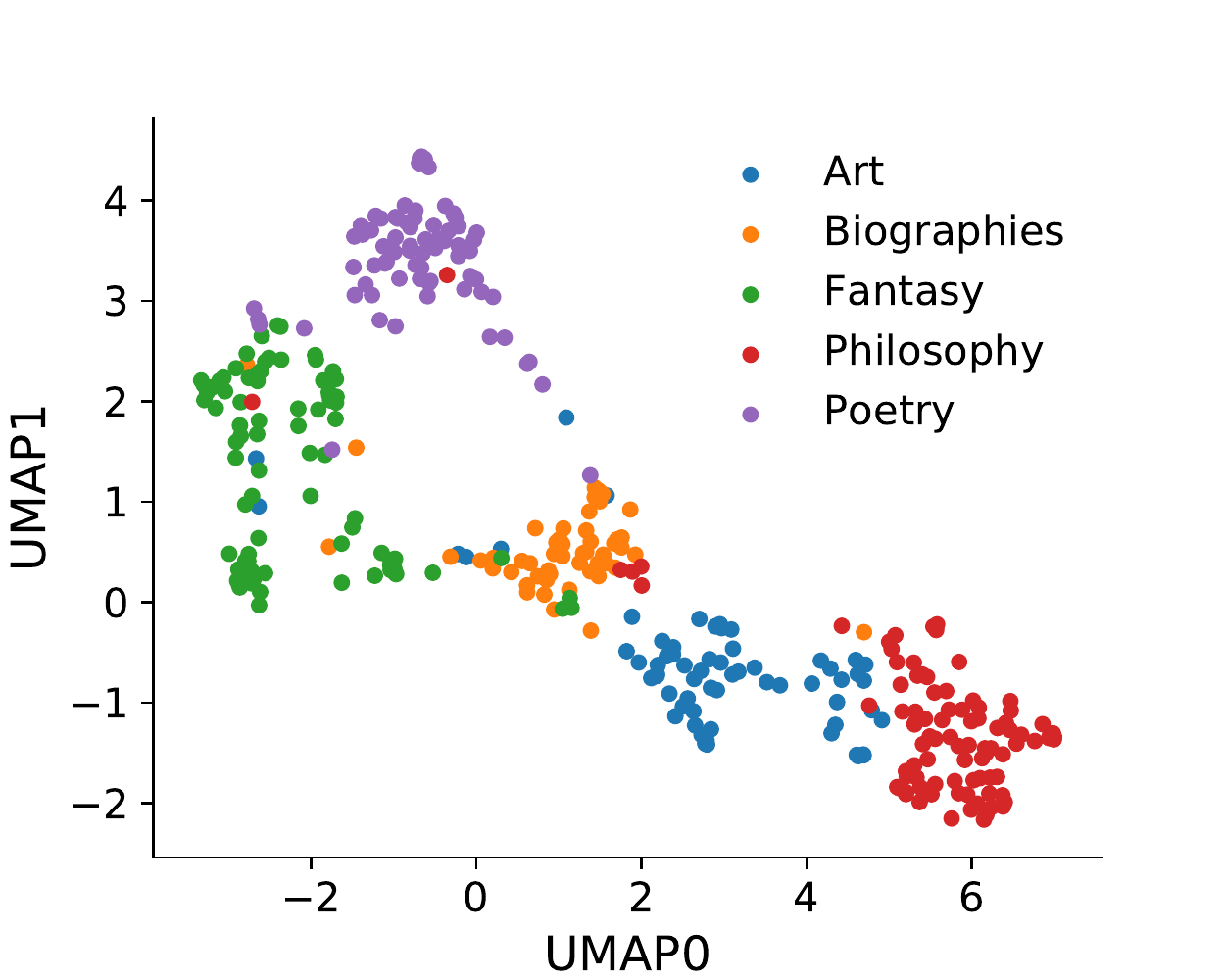}
 \caption{$2$-dimensional embedding shows clustering of books from the same bookshelf.
 Approximate visualization of the pair-wise distances between 370 PG books using UMAP (for details, see Section~\ref{sec.mat.met}).
 Each dot corresponds to one book colored according to the bookshelf membership.
 }
 \label{fig.umap}
\end{figure}

\paragraph*{Labels.} We select the 370 books tagged with one of the following bookshelf labels: Art, Biographies, Fantasy, Philosophy and Poetry. 
After calculating  distances $D_{i,j}$ between all pairs of books, in Fig.~\ref{fig.umap} we show an approximate 2-dimensional embedding (UMAP, see \cite{McInnes2018-uy} and Methods for details) in order to visualize which books are more similar to each other. 
Indeed, we find that books from the same bookshelf tend to cluster together and are well-separated from books belonging to other bookshelves.
This example demonstrates the usefulness of the bookshelf labels and that they reflect the topical variability encoded in the statistics of word frequencies.

\begin{figure}
 \centering
 \includegraphics[width=1\columnwidth]{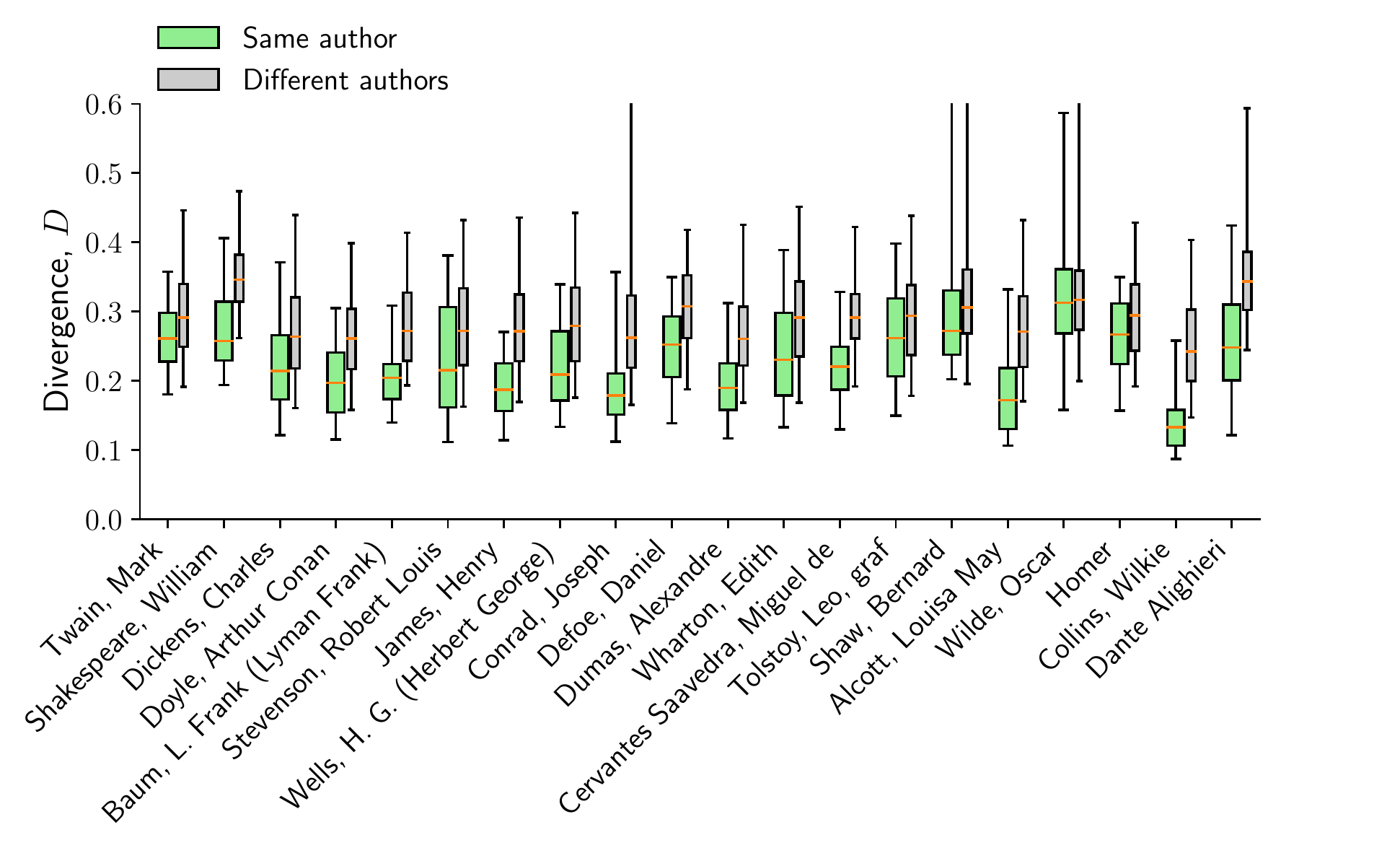}
 \caption{Distance between books from the same author is significantly smaller than distance between books from different authors.
 For each author, the boxplots shows the $5,25,50,75,95$-percentile of the distribution of distances from $1,000$ pairs of books from the same author (green) and to a different author (gray).
 }
 \label{fig.author}
\end{figure}

\paragraph*{Authors.} We select all books from the 20 most prolific authors~\footnote{We select these 20 authors from the authors of the 100 most downloaded books in order to avoid authors such as ``Anonymous''}.
For each author, we draw $1,000$ pairs of books $(i,j)$ from the same author and compare the distance $D_{i,j}$ with $1000$ pairs $(i,j')$ where $j'$ comes from a different author.
We observe that the distance between books from the same author is consistently smaller than for 2 books from different authors -- not only in terms of the median, but also in terms of a much smaller spread in the values of $D_{i,j}$ (Fig.~\ref{fig.author}).
This consistent variability across authors suggest the potential applicability in the study of stylistic differences, such as in problems of authorship attribution~\cite{Juola2008-mb,Stamatatos2009-uz}.

\begin{figure}
 \centering
 \includegraphics[width=0.9\columnwidth]{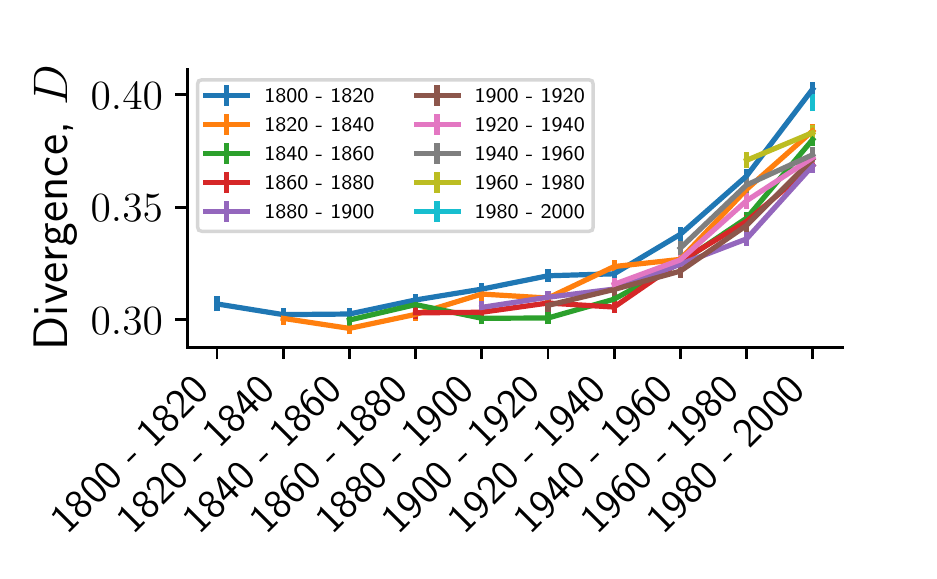}
 \caption{
 Distance between books increases with their time separation.
 Average and standard error of the distance between $1,000$ pairs of books, where the 2 books in each pair is drawn from 2 different 20-year time intervals.
 We fix the first interval and continuously increase the second time interval.
 }
 \label{fig.divergence-time}
\end{figure}

\paragraph*{Time.} We compare the distance $D_{i,j}$ between pairs of books $i, j$ taken each from a 20-year time period $t_i, t_j \in \{ 1800-1820, \ldots, 1980-2000 \}$. 
In Fig.~\ref{fig.divergence-time}, we show the distance between two time windows $D_{t_i,t_j}$ by averaging over each $1,000$ pairs of books.
We observe that the average distance increases with increasing separation between the time periods.
However, we emphasize that we only observe a substantial increase in $D_{t_i,t_j}$ for large separation between $t_i$ and $t_j$ and later time periods (after $1900$).
This could be caused by the rough approximation of the publication year and a potential change in composition of the SPGC after 1930 due to copyright laws.
This suggests the limited applicability of PG books for diachronic studies without further filtering (such as subject/bookshelf).

\section{Discussion}
We have presented the Standardized Project Gutenberg Corpus (SPGC), a decentralized, dynamic multilingual corpus containing more than $50,000$ books from more than $20$ languages.
Combining the textual data with metadata from two different sources we provided not only a characterization of the content of the full PG data but also showed three examples for resolving language variability across subject categories, authors, and time.
As part of this work, we provide the code for all pre-processing steps necessary to obtain a full local copy of the PG data.
We also provide a static or `frozen' version of the corpus, SPGC-2018-07-18, which ensures reproducibility of our results and can be downloaded at \href{https://doi.org/10.5281/zenodo.2422560}{https://doi.org/10.5281/zenodo.2422560}.

We believe that the SPGC will be a first step towards a more rigorous approach for using Project Gutenberg as a scientific resource.
A detailed account of each step in the pre-processing, accompanied by the corresponding code, are necessary requirements that will help ensure replicability in the statistical analysis of language and quantitative linguistics, especially in view of the crisis in reproducibility and replicability reported in other fields~\cite{Ioannidis2005-no,Open_Science_Collaboration2015-zj,Camerer2018-qt}.
From a practical point of view, the availability of this resource in terms of the code and the frozen dataset will certainly allow for an easier access to PG data, in turn facilitating the usage of larger and less biased datasets increasing the statistical power of future analysis.

We want to highlight the challenges of the SPGC in particular and PG in general, some of which can hopefully be addressed in the future.
First, the PG data only contains copyright-free books. 
As a result the number of books published after 1930's is comparably small.
However, in the future this can be expected to change as copyright for many books will run out and the PG data is continuously growing.
This highlights the importance of using a dynamic corpus model that will by default incorporate all new books when the corpus is generated for the first time.
Second, the annotation about the books is incomplete, and some books might be duplicated.
For example, the metadata lacks the exact date when a book was published --hindering the usage of the PG data for diachronic studies. Different editions of the same book might have been given a different PG identifier, and so they are all included in PG and thus in SPGC. 
Third, the composition of SPGC is heterogeneous, mixing different genres. However, the availability of document labels from the bookshelf metadata allows for systematic control of corpus composition. For example, it is easy to restrict to or exclude individual genres such as ``Poetry''. 

From a practical perspective, the SPGC has a strong potential to become a complementary resource in applications ranging from  computational linguistics to machine learning.
We emphasize that the SPGC contains thousands of annotated books in multiple languages even beyond the Indo-European language family.
There is an increasing interest in quantitative linguistics in studies beyond the English language.
In the framework of culturomics, texts could be annotated and weighted by additional metadata, e.g. in terms of their `success' measure as the number of readers~\cite{Yucesoy2018-yf} or number of PG downloads. For example, it could be expected that the impact of Carroll's ``Alice in Wonderland'' is larger than that of the ``CIA Factbook 1990''.
Furthermore, with an increase in the quality of the metadata, the identification of the same book in different languages might allow for the construction of high-quality parallel corpora used in, e.g. translation tasks.
Finally, in applications of Information Retrieval, metadata labels can be used to evaluate machine learning algorithms for classification and prediction.
These and other applications might require additional pre-processing steps (such as stemming) but  which could make use of SPGC as a starting point.

In summary, we believe that the SPGC is a first step towards a better usage of PG in scientific studies, and hope that its decentralized, dynamic and multi-lingual nature will lead to further collaborative interdisciplinary approaches to quantitative linguistics.

\section{Materials and Methods}
\label{sec.mat.met}

\subsection*{Pre-processing}
\paragraph*{Running the pipeline.} The simplest way to get a local copy of the PG database, with standardized, homogeneous pre-processing, is to clone the git repository
\begin{verbatim}
$ git clone git@github.com:pgcorpus/gutenberg.git
\end{verbatim}
and enter the newly created directory. To get the data, simply run:
\begin{verbatim}
$ python get_data.py
\end{verbatim}
This will download all available PG books in a hidden '.mirror' folder and symlink in the more convenient `data/raw' folder. To actually process the data, that is, to remove boiler-plate text, tokenize texts, filter and lowercase tokens, and count word type occurrence, it suffices to run 
\begin{verbatim}
$ python process_data.py
\end{verbatim}
which will fill in the rest of directories inside 'data' .

\paragraph*{Database synchronization.} We use `rsync' to keep an updated local mirror of \url{aleph.gutenberg.org::gutenberg}.

\paragraph*{Removal of duplicated entries.} Some PG book identifiers are stored in more than one location in PG's server. In these cases, we only keep the latest, most up-to-date version. We do not remove duplicated entries on the basis of book metadata or content.
\paragraph*{Removal of boiler-plate text.} To eliminate boiler-plate text that does not pertain to the books themselves, we use a list of known markers (code adapted from \url{https://github.com/c-w/gutenberg/blob/master/gutenberg/cleanup/strip_headers.py}).
\paragraph*{Text tokenization} Texts are tokenized via NLPToolkit~\cite{Loper2002-vq}. In particular, we set the `TreebankWordTokenizer` as the default choice, but this can be changed at will. Tokenization works better when the language of the text being analyzed is known. 
Since the metadata contains a language field for every downloaded book, we pass this information onto the tokenizer.
If the language field contains multiple languages ($\approx 0.3\%$ of the books), we use the first entry.
\paragraph*{Lower-casing and token filtering.} 
We only keep tokens composed entirely of alphabetic characters (including accented characters), removing those that contain digits or other symbols. Notice that this is done after tokenization, which correctly handles apostrophes, hyphens, etc. 
This constitutes a conservative approach to avoid artifacts, for example originating from the digitization process. 
While one might want to also include tokens with numeric characters in order to keep mentions of, e.g. years, the downside of this approach would be a substantial number of occurrences of undesirable page and chapter numbers.
However, we note that the modularized filtering can be easily customized (and extended) to incorporate also other aspects such as stemming as there is no one-size-fits all solution to each individual application.
Furthermore, all tokens are lower-cased. While this has undesired consequences in some cases (e.g. some proper nouns can be confounded with unrelated common nouns after being lower-cased), it is a simple and effective way of handling words capitalized after full stop or in dialogues, which would otherwise be (incorrectly) considered different words from their lowercase standard form.

\subsection*{2-dimensional embedding}
We use \emph{Uniform Manifold Approximation and Projection} (UMAP, \cite{McInnes2018-uy}) for visualization purposes in Fig.~\ref{fig.umap}. UMAP is a manifold-learning technique with strong mathematical foundations, see \cite{McInnes2018-uy} for details. We used normalized counts data as the input data, with word types playing the role of dimensions (features) and books playing the role of points (samples). Distance between points was computed using the Jensen-Shannon divergence~\cite{Gerlach2016-ld}. The end result is the 2-dimensional projection shown in Fig.~\ref{fig.umap}. Notice that subject labels were \emph{not} passed to UMAP, so the the observed clustering demonstrates that the statistics of word frequencies encode and reflect the manually-assigned labels.

\subsection*{Data availability.}
The code that we make available as part of this work allows to download and process all available Project Gutenberg books, facilitating the task of keeping an up-to-date and homogeneously processed dataset of a continuously growing resource. In fact, new books are added to Project Gutenberg daily. An unwanted consequence of this feature, however, is that two local versions of the SPGC might differ if they were last updated on different dates. 
To facilitate and promote reproducibility of our results and possible subsequent analysis, we provide a 'frozen' copy of the SPGC, last updated on 2018-07-18, containing 55,905 PG books. All statistics and figures reported on this manuscript are based on this version of the data. 
This data is available at \href{https://doi.org/10.5281/zenodo.2422560}{https://doi.org/10.5281/zenodo.2422560}.

\paragraph*{Requirements}
The 'frozen' dataset of all 55,905 books and all levels of granularity has a size of 65GB. 
However, focusing only on the one-gram counts requires only 3.6GB.
Running the pre-processing pipeline of the 'frozen' data took 8 hours (without parallelization) on an CPU running at 3.40GHz. 

\subsection*{Code availability.}
Python 3.6 code to download and pre-process all PG books can be obtained at \url{https://github.com/pgcorpus/gutenberg}, while python-based jupyter notebooks that reproduce the results of this manuscript can be obtained at \url{https://github.com/pgcorpus/gutenberg-analysis}.

\section*{Acknowledgments}
We warmly thank Ramon Ferrer-i-Cancho for encouraging us to pursue this project, as well as the organizers of the ``Statistics of Languages: Theories and Experiments'' meeting (Warsaw, July 2017).

\bibliographystyle{vancouver}
\bibliography{biblio_fran,biblio_martin}

\begin{thebibliography}{10}

\bibitem{Altmann2016-nl}
Altmann EG, Gerlach M.
\newblock Statistical laws in linguistics.
\newblock In: Degli~Esposti M, Altmann EG, Pachet F, editors. Creativity and
  Universality in Language. Springer; 2016. p. 7--26.

\bibitem{Ferrer_i_Cancho2003-he}
Ferrer~i Cancho R, Sol{\'e} RV.
\newblock Least effort and the origins of scaling in human language.
\newblock Proceedings of the National Academy of Sciences.
  2003;100(3):788--791.

\bibitem{Petersen2012-of}
Petersen AM, Tenenbaum JN, Havlin S, Stanley HE, Perc M.
\newblock Languages cool as they expand: Allometric scaling and the decreasing
  need for new words.
\newblock Scientific reports. 2012;2:943.

\bibitem{Tria2014-li}
Tria F, Loreto V, Servedio VDP, Strogatz SH.
\newblock The dynamics of correlated novelties.
\newblock Scientific reports. 2014;4:5890.

\bibitem{Corominas-Murtra2015-mp}
Corominas-Murtra B, Hanel R, Thurner S.
\newblock Understanding scaling through history-dependent processes with
  collapsing sample space.
\newblock Proceedings of the National Academy of Sciences. 2015;112:5348--5353.

\bibitem{Font-Clos2015-qy}
Font-Clos F, Corral A.
\newblock {Log-{log}} {c}onvexity of {{t}ype-{t}oken} {g}rowth in {Z}ipf's
  {s}ystems.
\newblock Phys Rev Lett. 2015;114:238701.

\bibitem{Cocho2015-hu}
Cocho G, Flores J, Gershenson C, Pineda C, S{\'a}nchez S.
\newblock Rank Diversity of Languages: Generic Behavior in Computational
  Linguistics.
\newblock PLoS One. 2015 Apr;10(4):e0121898.

\bibitem{Lippi2018-ua}
Lippi M, Montemurro MA, Esposti MD, Cristadoro G.
\newblock Natural Language Statistical Features of {LSTM-generated} Texts.
\newblock arXiv:180404087;.

\bibitem{Mazzolini2018-po}
Mazzolini A, Gherardi M, Caselle M, Cosentino~Lagomarsino M, Osella M.
\newblock Statistics of shared components in complex component systems.
\newblock Physical Review X. 2018;8:021023.

\bibitem{Dorogovtsev2001-yi}
Dorogovtsev SN, Mendes JF.
\newblock Language as an evolving word web.
\newblock Proc R Soc B. 2001;268:2603--2606.

\bibitem{Sole2010-hi}
Sol{\'e} RV, Corominas-Murtra B, Valverde S, Steels L.
\newblock Language networks: Their structure, function, and evolution.
\newblock Complexity. 2010;15(6):20--26.

\bibitem{Amancio2011-ww}
Amancio DR, Altmann EG, Oliveira ON, Costa LDF.
\newblock Comparing intermittency and network measurements of words and their
  dependence on authorship.
\newblock New Journal of Physics. 2011;13:123024.

\bibitem{Choudhury2010-bk}
Choudhury M, Chatterjee D, Mukherjee A.
\newblock Global topology of word co-occurrence networks: Beyond the two-regime
  power-law.
\newblock In: Proceedings of the 23rd International Conference on Computational
  Linguistics (Coling 2010); 2010. p. 162--170.

\bibitem{Cong2014-ia}
Cong J, Liu H.
\newblock Approaching human language with complex networks.
\newblock Physics of Life Reviews. 2014;11:598--618.

\bibitem{Bochkarev2014-kb}
Bochkarev V, Solovyev V, Wichmann S.
\newblock Universals versus historical contingencies in lexical evolution.
\newblock Journal of the Royal Society Interface. 2014;11:20140841.

\bibitem{Ghanbarnejad2014-qx}
Ghanbarnejad F, Gerlach M, Miotto JM, Altmann EG.
\newblock Extracting information from S-curves of language change.
\newblock Journal of the Royal Society Interface. 2014;11:20141044.

\bibitem{Feltgen2017-wy}
Feltgen Q, Fagard B, Nadal JP.
\newblock Frequency patterns of semantic change: Corpus-based evidence of a
  near-critical dynamics in language change.
\newblock Royal Society Open Science. 2017;4:170830.

\bibitem{Goncalves2018-yf}
Gon{\c c}alves B, Loureiro-Porto L, Ramasco JJ, S{\'a}nchez D.
\newblock Mapping the Americanization of English in space and time.
\newblock PLoS ONE. 2018;13:e0197741.

\bibitem{Amato2018-cu}
Amato R, Lacasa L, D{\'\i}az-Guilera A, Baronchelli A.
\newblock The dynamics of norm change in the cultural evolution of language.
\newblock Proc Natl Acad Sci U S A. 2018;115:8260--8265.

\bibitem{Karjus2018-ne}
Karjus A, Blythe RA, Kirby S, Smith K.
\newblock Challenges in detecting evolutionary forces in language change using
  diachronic corpora.
\newblock arXiv:181101275. 2018;.

\bibitem{Montemurro2010-mi}
Montemurro MA, Zanette DH.
\newblock Towards the quantification of the semantic information encoded in
  written language.
\newblock Advances in Complex Systems. 2010;13:135.

\bibitem{Takahira2016-mn}
Takahira R, Tanaka-Ishii K, D{e}bowski {\L}.
\newblock Entropy rate estimates for natural {language---A} new extrapolation
  of compressed {large-scale} corpora.
\newblock Entropy. 2016;18:364.

\bibitem{Febres2017-nt}
Febres G, Jaff{\'e} K.
\newblock Quantifying structure differences in literature using symbolic
  diversity and entropy criteria.
\newblock Journal of Quantitative Linguistics. 2017;24:16--53.

\bibitem{Bentz2017-ux}
Bentz C, Alikaniotis D, Cysouw M, Ferrer-i Cancho R.
\newblock The entropy of {words---Learnability} and expressivity across more
  than 1000 languages.
\newblock Entropy. 2017;19:275.

\bibitem{Ferrer_i_Cancho2004-eu}
Ferrer~i Cancho R, Sol{\'e} R, K{\"o}hler R.
\newblock Patterns in syntactic dependency networks.
\newblock Physical Review E. 2004;69:51915.

\bibitem{Kulig2016-hy}
Kulig A, Kwapie{\'n} J, Stanisz T, Dro{\. z}d{\. z} S.
\newblock In narrative texts punctuation marks obey the same statistics as
  words.
\newblock Inf Sci. 2016;375:98--113.

\bibitem{Michel2011-ya}
Michel JB, Shen YK, Aiden AP, Veres A, Gray MK, Team TGB, et~al.
\newblock Quantitative analysis of culture using millions of digitized books.
\newblock Science. 2011;331(6014):176--182.

\bibitem{Masucci2011-db}
Masucci AP, Kalampokis A, Egu{\'\i}luz VM, Hern{\'a}ndez-Garc{\'\i}a E.
\newblock Wikipedia information flow analysis reveals the scale-free
  architecture of the semantic space.
\newblock PLoS One. 2011;6:e17333.

\bibitem{Yasseri2012-lf}
Yasseri T, Kornai A, Kert{\'e}sz J.
\newblock A practical approach to language complexity: A Wikipedia case study.
\newblock PloS One. 2012;7:e48386.

\bibitem{Dodds2011-je}
Dodds PS, Harris KD, Kloumann IM, Bliss CA, Danforth CM.
\newblock Temporal patterns of happiness and information in a global social
  network: Hedonometrics and twitter.
\newblock PLoS One. 2011;6:e26752.

\bibitem{Morse-Gagne2011-pj}
Morse-Gagn{\'e} EE.
\newblock Culturomics: statistical traps muddy the data.
\newblock Science. 2011;332(6025):35.

\bibitem{Pechenick2015-ov}
Pechenick EA, Danforth CM, Dodds PS.
\newblock Characterizing the Google Books corpus: Strong limits to inferences
  of {socio-cultural} and linguistic evolution.
\newblock PLoS One. 2015;10:e0137041.

\bibitem{Hart_undated-vb}
Hart M. Project Gutenberg; 1971.

\bibitem{Ebeling1994-nc}
Ebeling W, P{\"o}schel T.
\newblock Entropy and {long-range} correlations in literary English.
\newblock Europhys Lett. 1994;26:241--246.

\bibitem{Schurmann1996-sv}
Schurmann T, Grassberger P.
\newblock Entropy estimation of symbol sequences.
\newblock Chaos. 1996;6:414--427.

\bibitem{Baayen1996-oc}
Baayen RHH.
\newblock The effects of lexical specialization on the growth curve of the
  vocabulary.
\newblock Comput Linguist. 1996;22:455--480.

\bibitem{Altmann2012-bk}
Altmann EG, Cristadoro G, Esposti MD.
\newblock On the origin of long-range correlations in texts.
\newblock Proc Natl Acad Sci U S A. 2012;109:11582--11587.

\bibitem{Moreno-Sanchez2016-ty}
Moreno-S{\'a}nchez I, Font-Clos F, Corral {\'A}.
\newblock {Large-scale} analysis of Zipf's law in English texts.
\newblock PLoS One. 2016;11:e0147073.

\bibitem{Williams2015-mz}
Williams JR, Bagrow JP, Danforth CM, Dodds PS.
\newblock Text mixing shapes the anatomy of rank-frequency distributions.
\newblock Phys Rev E. 2015;91:052811.

\bibitem{Tria2018-ic}
Tria F, Loreto V, Servedio V.
\newblock Zipf's, Heaps' and Taylor's Laws are determined by the expansion into
  the adjacent possible.
\newblock Entropy. 2018;20:752.

\bibitem{Hughes2012-xs}
Hughes JM, Foti NJ, Krakauer DC, Rockmore DN.
\newblock Quantitative patterns of stylistic influence in the evolution of
  literature.
\newblock Proceedings of the National Academy of Sciences. 2012;109:7682--7686.

\bibitem{Reagan2016-yp}
Reagan AJ, Mitchell L, Kiley D, Danforth CM, Dodds PS.
\newblock The emotional arcs of stories are dominated by six basic shapes.
\newblock EPJ Data Science. 2016;5:31.

\bibitem{Dua:2017}
Dheeru D, Karra~Taniskidou E. {UCI} Machine Learning Repository; 2017.
\newblock Available from: \url{http://archive.ics.uci.edu/ml}.

\bibitem{COCA-corpus}
Davies M. The Corpus of Contemporary American English (COCA): 560 million
  words, 1990-present.; 2008.
\newblock Available from: \url{https://corpus.byu.edu/coca/}.

\bibitem{Leech1993-ub}
Leech G.
\newblock 100 million words of English.
\newblock English Today. 1993;9:9--15.
\newblock Available from: \url{http://www.natcorp.ox.ac.uk}.

\bibitem{Cattuto2007-uo}
Cattuto C, Loreto V, Pietronero L.
\newblock Semiotic dynamics and collaborative tagging.
\newblock Proc Natl Acad Sci U S A. 2007;104:1461--1464.

\bibitem{Loper2002-vq}
Loper E, Bird S.
\newblock {NLTK}: The Natural Language Toolkit.
\newblock In: Proceedings of the {ACL-02} Workshop on Effective Tools and
  Methodologies for Teaching Natural Language Processing and Computational
  Linguistics - Volume 1. ETMTNLP '02. Stroudsburg, PA, USA: Association for
  Computational Linguistics; 2002. p. 63--70.

\bibitem{Manning2008-sf}
Manning CD, Raghavan P, Sch{\"u}tze H.
\newblock Introduction to Information Retrieval.
\newblock Cambridge University Press; 2008.

\bibitem{Gerlach2016-ld}
Gerlach M, Font-Clos F, Altmann EG.
\newblock Similarity of symbol frequency distributions with heavy tails.
\newblock Phys Rev X. 2016;6:021009.

\bibitem{McInnes2018-uy}
McInnes L, Healy J.
\newblock {UMAP}: Uniform manifold approximation and projection for dimension
  reduction.
\newblock arXiv:180203426. 2018;.

\bibitem{Note1}
;.
\newblock We select these 20 authors from the authors of the 100 most
  downloaded books in order to avoid authors such as ``Anonymous''.

\bibitem{Juola2008-mb}
Juola P.
\newblock Authorship attribution.
\newblock Foundations and Trends\textregistered{} in Information Retrieval.
  2008;1:233--334.

\bibitem{Stamatatos2009-uz}
Stamatatos E.
\newblock A survey of modern authorship attribution methods.
\newblock Journal of the American Society for Information Science and
  Technology. 2009;60:538--556.

\bibitem{Ioannidis2005-no}
Ioannidis JPA.
\newblock Why most published research findings are false.
\newblock PLoS Medicine. 2005;2:e124.

\bibitem{Open_Science_Collaboration2015-zj}
{Open Science Collaboration}.
\newblock Estimating the reproducibility of psychological science.
\newblock Science. 2015;349:aac4716.

\bibitem{Camerer2018-qt}
Camerer CF, Dreber A, Holzmeister F, Ho TH, Huber J, Johannesson M, et~al.
\newblock Evaluating the replicability of social science experiments in Nature
  and Science between 2010 and 2015.
\newblock Nature Human Behaviour. 2018;2:637--644.

\bibitem{Yucesoy2018-yf}
Yucesoy B, Wang X, Huang J, Barab{\'a}si AL.
\newblock Success in books: A big data approach to bestsellers.
\newblock EPJ Data Science. 2018;7:7.

\end{thebibliography}

\end{document}